%% file: iclr2022_conference.tex
\documentclass{article} 
\usepackage{iclr2022_conference,times}

\input{math_commands.tex}

\usepackage{hyperref}
\usepackage{url}
\usepackage{graphicx}
\usepackage{caption}
\usepackage{subcaption}
\usepackage{wrapfig}

\title{Efficiently Evolving Swarm Behaviors Using Grammatical Evolution With PPA-style Behavior Trees}

\author{Aadesh Neupane \& Michael A. Goodrich  
\\
Department of Computer Science\\
Brigham Young University\\
Provo, UT 84602, USA \\
\texttt{adeshnpn@byu.edu, mike@cs.byu.edu} \\
}

\usepackage{listings}
\usepackage[ruled,vlined]{algorithm2e}
\usepackage[nounderscore]{syntax}

\newcommand{\br}[0]{BeTr }

\def\squash{
      \setlength{\itemsep}{2pt}
      \setlength{\partopsep}{0pt}
      \setlength{\parskip}{0pt}
      \setlength{\parindent}{0pt}
      \setlength{\topsep}{0pt}
      \setlength{\partopsep}{0pt}}

\usepackage{etoolbox} 
\makeatletter
\def\gr@implnumbereditem<#1> #2 {%
  \stepcounter{grammarline}%
  \sbox\z@{\hskip\labelsep\grammarlabel{#1}{#2}}
  \strut\@@par%
  \vskip-\parskip%
  \vskip-\baselineskip%
  \hrule\@height\z@\@depth\z@\relax%
  \item[%
    \rlap{\hskip\dimexpr\linewidth+\grammarindent\relax 
          \llap{(\thegrammarline)}}%
    \unhbox\z@]%
  \catcode`\<\active%
}
\let\numberedgrammar\grammar

\pretocmd\numberedgrammar{\setcounter{grammarline}{0}}{}{}
\patchcmd\numberedgrammar
  {\gr@implitem}
  {\gr@implnumbereditem}
  {}{}
\patchcmd\numberedgrammar
  {\def\alt{\\\llap{\textbar\quad}}}
  {\let\alt\alt@num}
  {}{}

\def\alt@num{\\\relax
  \stepcounter{grammarline}%
  \rlap{\hskip\dimexpr\linewidth-\labelwidth+\grammarindent-\labelsep\relax
        \llap{(\thegrammarline)}}
  \llap{\textbar\quad}}

\newcounter{grammarline}
\makeatother

\newcommand{\BibTeX}{\rm B\kern-.05em{\sc i\kern-.025em b}\kern-.08em\TeX}

\begin{document}

\maketitle

\begin{abstract}
Evolving swarm behaviors with artificial agents is computationally expensive and challenging. Because reward structures are often sparse in swarm problems, only a few simulations among hundreds evolve successful swarm behaviors. Additionally, swarm evolutionary algorithms typically rely on ad hoc fitness structures, and novel fitness functions need to be designed for each swarm task. This paper evolves swarm behaviors by systematically combining Postcondition-Precondition-Action (PPA) canonical Behavior Trees (BT) with a Grammatical Evolution. The PPA structure replaces ad hoc reward structures with systematic postcondition checks, which allows a common grammar to learn solutions to different tasks using only environmental cues and BT feedback. The static performance of learned behaviors is poor because no agent learns all necessary subtasks, but performance while evolving is excellent because agents can quickly change behaviors in new contexts.  The evolving algorithm succeeded in 75\% of learning trials for both foraging and nest maintenance tasks, an eight-fold improvement over prior work. 

\end{abstract}

\section{Introduction}

Bio-inspired models have produced efficient algorithms in various domains~\citep{karaboga2007powerful,dorigo2006ant}. The potential benefits of bio-inspired algorithms are limited by the cumbersome task of observing animal behavior and creating mathematical models to describe both individual and collective behaviors~\citep{sumpter2003modelling,gordon2010ant}. A promising alternative is to design novel swarm behaviors using evolutionary algorithms.  
 
Evolving swarm behaviors in non-episodic setting requires group-level objectives and rewards to be divided into individual objectives and rewards, which is a form of the \textit{credit assignment problem} (CAP)~\citep{sutton1984temporal}. Existing swarm evolution algorithms address CAP by designing ad hoc fitness functions that give the artificial agents sufficient feedback to solve a particular task~\citep{ferrante2013geswarm,neupane2018geese}. Unfortunately, designing an ad hoc fitness function for each task requires expert knowledge, and is subject to human biases~\citep{nelson2009fitness}. Also, evolving swarm behaviors require the designer to choose and aggregate various controllers, evolutionary algorithms, and fitness functions. In many cases, only a few combinations are viable to evolve collective behaviors, and the success rate is low among those viable combinations.

A multi-agent Grammatical Evolution (GE)~\citep{o2001grammatical} algorithm called GEESE evolved colony-level foraging behaviors, outperforming conventional GE and hand-coded solutions on a foraging task~\citep{neupane2018geese}. GEESE's controllers were Finite State Machines (FSMs) and the group fitness was the total food collected, but GEESE could only evolve foraging behaviors. GEESE-BT~\citep{neupane2019learning} used a more expressive grammar that could evolve both foraging and cooperative transport behaviors. GEESE-BT's controllers were behavior trees (BTs), and GEESE-BT used ad hoc fitness functions from the theory of intrinsic motivators. Unfortunately, the ad hoc reward structures were not sufficient to evolve successful behaviors with high success rate and required a new fitness function for each new task. 

This paper presents the \br-GEESE (``better geese") algorithm, which produces successful swarm behaviors while learning even though no single agent has all necessary behaviors at any given point during the evolution.  \br-GEESE improves GEESE-BT in three significant ways: (i)~primitive behaviors use Postcondition-Precondition-Action (PPA) structures~\citep{sprague2018improving}; (ii)~
the grammar used PPA-style BT programs; and (iii)~the task-specific ad hoc rewards were replaced with BT execution node status, enabling direct feedback from PPA nodes.  The evolving algorithm succeeds in 75\% of learning trials for two tasks (foraging and nest maintenance), an eight-fold improvement over GEESE-BT. 

Two limitations are noteworthy. First, swarm success requires agents to performing ongoing evolution. This is problematic if agents experience catastrophic failures while learning, such as when a physical robot crashes in the world. Second, the algorithm is only applied to multiagent tasks that are \textit{divisible} and \textit{additive} tasks, meaning that the swarm's task can be broken into subtasks achievable by individual programs~\citep{steiner1972group}.




\section{Related Work}


Early work in evolving robot controllers include (a)~staged evolution of a complex motor pattern generator to control a walking robot~\citep{lewis1992genetic} and (b)~evolving  neural-network-based control architectures for visually guiding robots~\citep{cli1993evolving}.  Neural networks can directly map robot inputs to outputs~\citep{lewis1996neural,trianni2003evolving}, but tend to be data-hungry and not transparent~\citep{fan2021interpretability}. \citet{kriesel2008beanbag} applied evolution to swarm systems and demonstrated that simple evolved individuals can produce effective swarm behaviors. 



Evolutionary robotics algorithms differ from each other in the choice of the evolutionary algorithm and the choice of the agent's controller. Individual robot controllers have been designed using Neural Networks (NN), Finite state machines (FSM) and hierarchical FSMs (HFSM)~\citep{petrovic2008evolving,pinter2012towards,konig2009decentralized,brooks1986robust,valmari1996state}. Behavior Trees (BT) often provide more readable, scalable, modular, and reactive structures than HFSMs and NNs~\citep{colledanchise2017behavior}.

\citet{duarte2016evolution} evolved swarm behaviors for physical robots and demonstrated scalability, flexibility, and resilience; results were shown for swarm-level dispersion, homing, clustering, and monitoring.  GE and state machine-type controllers have been combined to evolve swarm behaviors~\citep{neupane2018geese,neupane2019designing,ferrante2013geswarm}.  Even though GE exploits prior knowledge in the form of grammars to learn better solutions much faster than conventional genetic programs~\citep{o2001grammatical}, results depend on the design of ad hoc reward structures.

The lateral genetic transfer of controllers in \br-GEESE is bio-inspired. Prior work claims that complex behaviors rarely evolve solely from crossover and mutation alone but require endosymbiosis or horizontal transfer~\citep{jablonka2014evolution,lane2015vital,quammen2018tangled}. With the horizontal transfer, evolving single agents with sparse and delayed reward is more computationally efficient than evolving complex controllers~\citep{lee1999evolving,engebraaten2018evolving}.  



PPA structures have been widely used~\citep{fikes1971strips,knoblock1995planning}. \citet{sprague2018improving} used PPA structures to construct a modular, versatile, and robust control architecture for mission-critical autonomous underwater vehicles. The PPA structure ensured that the execution of BT followed goal fulfillment priorities.
~\citet{colledanchise2019towards} introduced a standard backward chaining algorithm to create a PPA structure automatically, and showed that the structure could skip actions that were already executed and only plan when the postconditions were not satisfied. ~\citet{ogren2020convergence} proved convergence guarantees for a particular PPA-BT structure, and ~\citet{parashar2021meta} presented a PPA layered strategy to transform a mission into decomposable tasks. BT-based collective behaviors have been used in non-evolutionary-based optimization. \citet{jonas2018behavior} used BTs to perform foraging and aggregation. \citet{kuckling2021automatic} extended that work with a set of behavioral modules within a predefined BT structure. The algorithm optimized swarm behaviors for foraging and marker aggregation tasks when agents could communicate.

\section{GEESE-BT Overview}
\label{sec:geese-bt-overview}

GEESE~\citep{neupane2018geese} and GEESE-BT~\citep{neupane2019learning} are multi-agent grammatical evolution (GE) algorithms that are similar to standard GE in terms of initialization, genetic operators, and genotype-to-phenotype mapping. Each agent is initialized with a fixed-length array of binary strings called the genome. The genome is then grouped into fixed consecutive blocks called a codon. Each codon block can be converted into a codon value. The conversion of the genome to a program with the help of BNF grammar is the genotype-to-phenotype mapping. The max-tree-depth parameter controls the number of unexpanded non-terminals in the current parse tree. 


 
GEESE agents uses three evolution steps: \textit{sense}, \textit{act}, and \textit{update}. During the \textit{sense} phase, agents exchange genome information with any (nearby) agents in the field of view. The willingness to transfer their gene is controlled by the INTERACTION\_PROB parameter. During the \textit{act} phase, an agent queries its storage pool to determine whether the pool size exceeds STORAGE\_THRESHOLD parameter. Higher thresholds indicate that interactions with more agents are needed before evolving. Like prior GE work~\citet{o2001grammatical}, if the threshold is exceeded, agents apply genetic operations to the gene pool in the order: a) selection, b) crossover, and c) mutation. Selection samples a subset of genotypes based on the fitness value to form a new population. Crossover combines two genomes to produce a new genome. Mutation randomly flips the bits of the genome. During the \textit{update} phase, an agent replaces its current genotype with a new geneotype if there is a new  genotype with higher fitness. Each evolution/learning time-step, all agents sense, act, and update.  
\begin{table}[h]
\begin{center}
\begin{tabular}{|c | c | c|} 
 \hline
 Parameters & GEESE & GEESE-BT\\
 \hline\hline
 Storage Threshold & NA & 7 \\ 
 \hline
 Interaction Probability & 0.8 & 0.85 \\ 
 \hline
 Parent-Selection & Tournament & Fitness + Truncation \\
 \hline
 Elite-size & 1 & N/A \\
 \hline
 Mutation Probability & 0.01 & 0.01 \\
 \hline
 Crossover Probability & 0.9 & 0.9 \\ 
 \hline
  Crossover & variable\_onepoint & variable\_onepoint \\ 
 \hline
  Genome-Selection & Tournament & Diversity \\ 
 \hline
  Number of Agents & 100 & 100 \\
  \hline
  Evolution Steps & 284 & 12000 \\
  \hline
  Max Tree Depth & 10 & 10 \\
  \hline
\end{tabular}
\caption{Evolution parameters used by GEESE and GEEST-BT.}\label{table:geese_param} 
\end{center}
\end{table}

GEESE-BT~\citep{neupane2019learning} replaced the state-machine controllers in GEESE with Behavior Trees (BT). Paraphrasing~\citet{neupane2019learning} for context, a BT is a directed rooted tree with internal \textit{control flow nodes} and leaf \textit{execution nodes}~\citep{colledanchise2018behavior}. BT~execution starts at the root node by generating \textit{ticks} at a fixed frequency. After each tick, a node returns \textit{running}, meaning that processing is ongoing, \textit{success}, meaning that the node's objective is achieved, or \textit{failure}, meaning neither running nor success. Control nodes include \textit{selectors}, which act as logical {\tt or}, \textit{parallel} and \textit{sequence} nodes, which act as logical {\tt and}, and \textit{condition} nodes, which act as logical {\tt if}.

\section{BeTr-GEESE}
BeTr-GEESE shares the core evolutionary stages (sense, act, and update) and genetic operations as GEESE and GEEST-BT, but has three significant improvements: a) PPA-based primitive behaviors, b) PPA-style BNF grammars, and c) a BT-induced fitness function. Since each of these changes impact swarm performance, each is experimentally evaluated in Sections~\ref{subsec:impppa}-\ref{subsec:fitness}, respectively.  BeTr-GEESE uses the same parameters as GEESE-BT (Table~\ref{table:geese_param}) to ensure unbiased comparison.

\subsection{Agent Lifecycle}
The BeTr-GEESE algorithm runs for a particular number of steps, which is defined before the start of the simulation\footnote{BeTr-GEESE learning is not episodic like some GE algorithms; there are no terminal states and replay.}. $n$~agents are initialized in a grid environment as described in Section~\ref{sec:experiment_conditions}, and a random genome of length 100 is created. The genome is transformed to a BT controller by the genotype-to-phenotype mapping process using the swarm BNF grammar described in appendix~\ref{app:bnfgrammar}. Each agent interacts with the environment using the BT controllers and updates its fitness value using \textit{diversity fitness} and an overall fitness function $\mathbf{A}_t$ described in Section~\ref{subsec:fitness}. (\textit{Diversity fitness} is the total number of unique behaviors nodes divided by the total behaviors defined in the grammar.)   Then the agents independently perform the sense, act, and update methods described in Section~\ref{sec:geese-bt-overview}.



\subsection{Simulation Parameters}
\label{sec:experiment_conditions}


A 100x100 grid environment was used with a hub of radius ten at the origin. The population had 100 learning agents. Agents moved with speed of 2 units per time step. Task performance is measured in terms of food or debris moved at the end of 12,000 learning steps. 12,000 learning steps was adopted from GEESE-BT~\citep{neupane2019learning}, and from empirical analysis it was observed that for all three task reported in GEESE-BT obtained best performance when the simulation was run for at least 12,000 steps.

The swarm simulation environment was created using the \textbf{Mesa} agent-based modeling framework~\citep{python-mesa-2020}. The simulator lacks the high fidelity physic engine, making it faster to run experiments with a large number of agents but greatly simplifies the robot-environment interactions. BT controllers for the agents were created using Behavior Tree framework \textbf{py\_trees}~\citep{py_trees} and \textbf{PonyGE2}~\citep{fenton2017ponyge2} was used to implement the Grammatical Evolution algorithms GEESE, GEESE-BT, and BeTr-GEESE.  All experiments were performed on a machine with an i9 CPU, 64 GB RAM running 16 parallel threads.

\subsection{Evaluation Metrics}
\br-GEESE is evaluated using two swarm tasks, \textit{Foraging} and \textit{Nest Maintenance}.
The \textit{Foraging} task requires agents to retrieve food from a source to a hub. A single foraging site of radius ten with multiple ``food'' objects is randomly placed at 30 units from the hub. The total food units is set to equal agent population. Task performance is the percentage of food at the hub.

The \textit{Nest Maintenance} task requires agents to move debris near the hub to a place outside a fixed boundary. Multiple ``debris'' objects are placed within ten units radius of the hub. The desired boundary is set at 30 units radius away from the origin. The total debris units at the hub is equal to the agent population. Task performance is the percentage of debris that is outside the boundary.


The simulation is labelled \textit{learning} when the agents are evolving controllers where the behaviors of the agent changes with time. The simulation is labelled \textit{test} when the agent controllers from a \textit{learning} simulation are transferred to a new simulation environment and the agents controllers remains static and do not evolve. For all the experiments reported in this paper, \textit{test} simulation does not differ drastically but only in the positions of sites and obstacles placed randomly at the runtime. Foraging is deemed \textit{successful} if more than $80\%$ food is collected, and nest maintenance is successful if more than $80\%$ debris is removed. \textit{Success rate} is defined as the ratio of the number of successful trials to the total number of simulations. \textit{Learning efficiency} is defined as foraging or maintenance percentage at the end of a learning simulation.

\subsection{Adding PPA to Primitive Behaviors}
\label{subsec:impppa}

\begin{figure}[htb]
  \centering %
	\includegraphics[width=.7\linewidth, keepaspectratio]{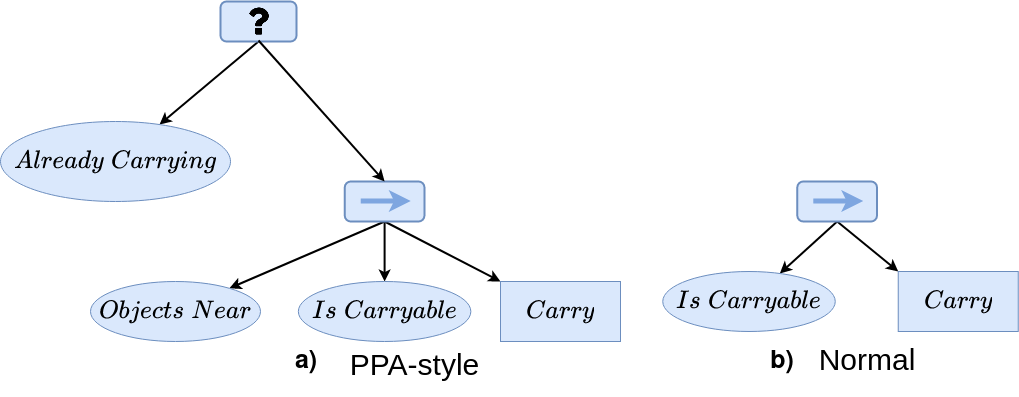}
    \caption{\footnotesize {BTs for \textit{CompositeSingleCarry} for a) PPA-style and b) nominal trees.}}
    \label{fig:ppa_carry} 
\end{figure}

Figure~\ref{fig:ppa_carry}(b) illustrates how GEESE-BT requires a precondition (\textit{IsCarryable}) to be satisfied {\tt and} the task (\textit{Carry}) successfully performed. The {\tt and} operator is implemented as a sequence BT node ($\rightarrow$). \br-GEESE uses the PPA structure in Figure~\ref{fig:ppa_carry}(a). The root selector node (represented as the ``{\tt ?}'') checks the postcondition (left branch, \textit{AlreadyCarrying}) and calls the sequence node in the right child only if the postcondition is not met. The right child checks preconditions  and implements the action.  The postcondition ensures that the planner does not need to re-execute when the goal has been met. The PPA structure shown is for the \textit{CompositeSingleCarry} behavior; see Appendix~\ref{section:PrimitiveBehaviors}.

\begin{figure}[htbp]
  \centering
  \begin{subfigure}[b]{0.48\textwidth}
	\includegraphics[width=0.99\linewidth, keepaspectratio]{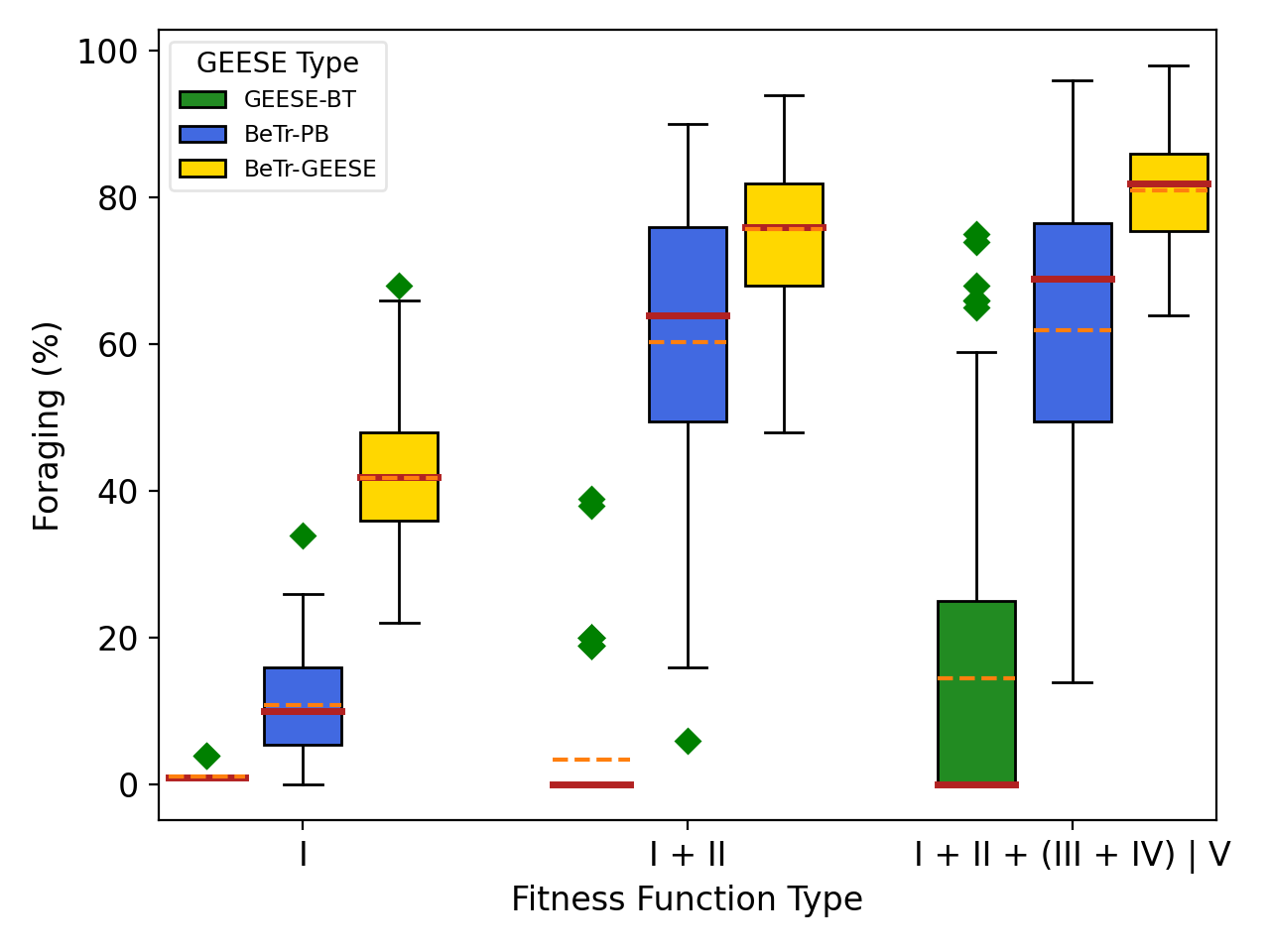}
    \caption{\footnotesize {Foraging experiments.}}
    \label{fig:BeTr-GEESE_Fitness_Comparision}
\end{subfigure}
\begin{subfigure}[b]{0.48\textwidth}
	\includegraphics[width=0.99\linewidth, keepaspectratio]{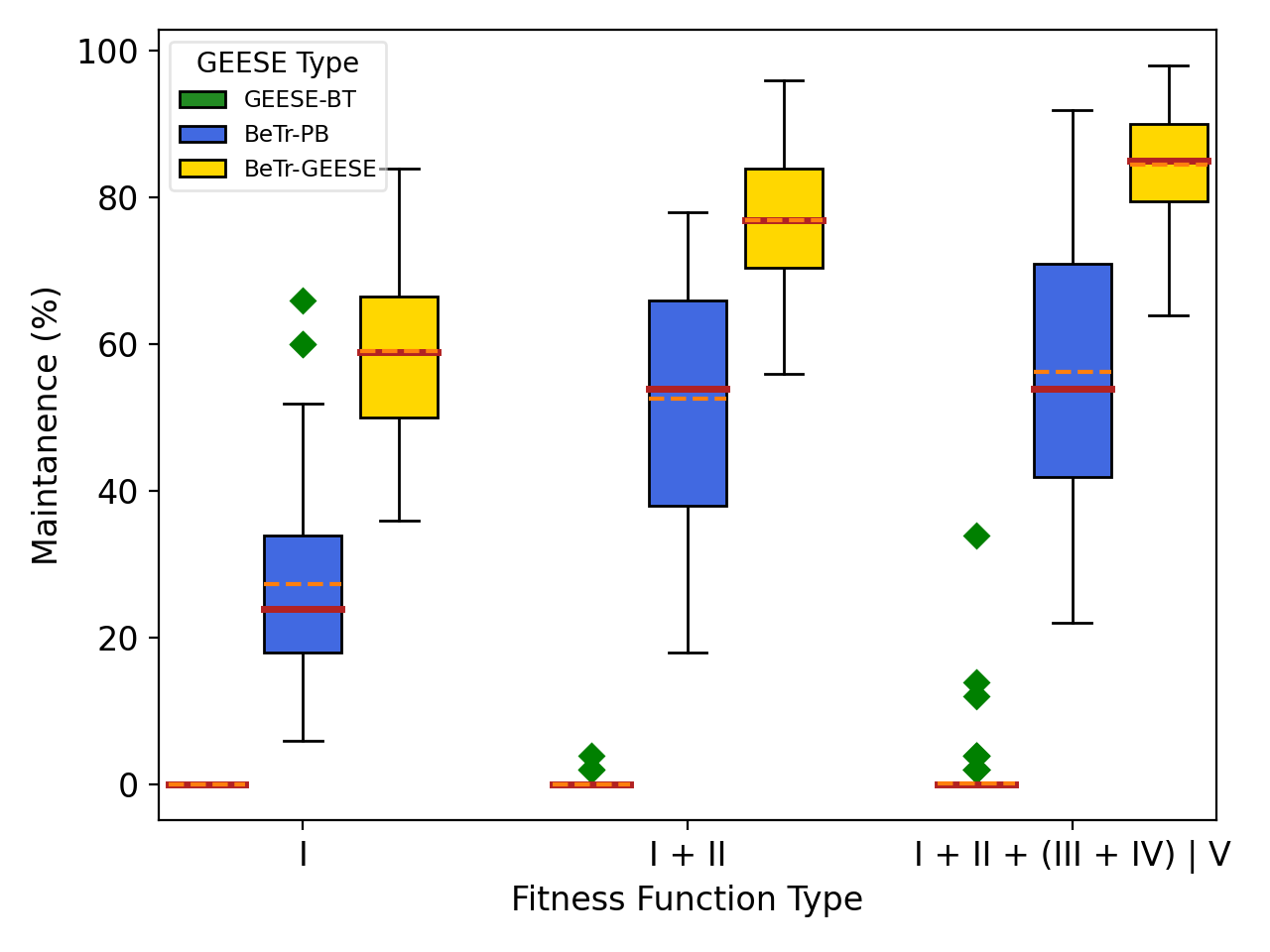}
    \caption{\footnotesize{Nest-maintenance experiments.}}
    \label{fig:BeTr-GEESE_Fitness_Comparision-Nest}
    \end{subfigure}
    \caption{Learning efficiency, measured by the percentage of food/debris transported to/from the hub with respect to variations in primitive behaviors, grammar, and fitness function. }
    \label{fig:BeTr-GEESE_vs_GEESE-BT}
\end{figure}

The green and blue boxes in Figures~\ref{fig:BeTr-GEESE_Fitness_Comparision}-\ref{fig:BeTr-GEESE_Fitness_Comparision-Nest} compare learning efficiency (over a range of fitness function conditions, described below) between the GEESE-BT primitive behaviors (PB) and the PPA-based primitive behaviors (\br-PB) for the two tasks.  The PPA structure improves performance across all fitness function conditions and for both tasks. 

\subsection{Adding PPA to the Grammar}
Appendix~\ref{app:bnfgrammar} presents the grammar used by \br-GEESE, which  is modified from the GEEST-BT grammar to use the PPA structure. The blue and yellow boxes in the leftmost and middle groups of Figures~\ref{fig:BeTr-GEESE_Fitness_Comparision}-\ref{fig:BeTr-GEESE_Fitness_Comparision-Nest} both use PPA-based PBs, but the yellow boxes use PPA-based grammars, respectively. 
Changing the grammar to use PPA structures further improves learning efficiency for both tasks.

\subsection{Replacing ad hoc Fitness with BT Status}
\label{subsec:fitness}

\textbf{Essential Fitness Elements.} There are various choices of the fitness function in evolutionary robotics based on the type of controllers and amount of priory task knowledge the designer has~\citep{nelson2009fitness}. GE requires a diverse genetic population, and both foraging and nest maintenance require exploring the world. Thus, two fitness elements are required: diversity (type~I) and exploration (type~II). The phenotype is an agent program learned from the grammar. For GEESE-BT and \br-GEESE, the program is a behavior tree.  \textit{Diversity fitness} is denoted by $D: \{ {\rm Phenotypes} \} \rightarrow \mathbb{R}$, promotes diversity in the genotype, and is defined as the total number of unique behavior nodes in the BT divided by the total possible behaviors defined in the grammar. \textit{Exploration fitness} is denoted by $E: \{{\rm Locations}\} \longrightarrow \mathbf{R}$, promotes visiting new locations, and is defined as the number of unique world locations visited by the agent.

\textbf{Ad hoc Fitness Elements.} GEESE-BT uses ad hoc fitness functions. Prospective fitness (type III) rewards ``intrinsic" actions like picking up, carrying, or dropping objects. Task-specific fitness (type IV) uses hand-tuned fitness functions designed to reward collective actions: total food at the hub and the total debris away from the hub, respectively. 

\textbf{BT Feedback Fitness.}  The BT feedback fitness (type~V) is denoted by $B: \{{\rm Behavior Tree Status}\} \longrightarrow \mathbf{R}$ and is defined as the sum of potstcondition, constraint, and selector node rewards. When a postcondition node status is \textit{success}, a subjectively chosen reward of~$+1$ indicates that some potentially useful condition in the world holds. A subjectively chosen reward of~$-2$ occurs when a constraint node status is \textit{failure}. A subjectively chosen reward of~$+1$ is returned when the root selector node status success, indicating that some sub-task has been accomplished somewhere in the BT. Based on~\cite{nelson2009fitness}, BT feedback (type~V) incorporates a low level of prior knowledge, whereas ad hoc fitness (type~III \& type~IV) incorporates high or excessive prior knowledge. BT feedback is superior to ad hoc fitness because it requires less prior knowledge, is general for all swarm tasks, and remains the same while the task changes. 

\textbf{Blending Fitness Types.} GEESE-BT blends types~I--IV to produce overall fitness. Since types~III and~IV are ad hoc, overall fitness is ad hoc.
By contrast, \br-GEESE rewards behaviors that promote genetic diversity, promote world exploration, observe or accomplish subtasks, or avoid constraint violations. Let $\mathbf{A}_t$ denote an agent's fitness, defined as the exponential blend 
$\mathbf{A}_t = \beta (\mathbf{A}_{t-1}) + (D + E_t + B_t)$, with $\beta$ empirically set to~$0.9$. Historical blending overcomes the temporal sparsity of diversity, exploration, and BT status. 




\textbf{Fitness Results.} The leftmost and middle group in Figures~\ref{fig:BeTr-GEESE_Fitness_Comparision}-\ref{fig:BeTr-GEESE_Fitness_Comparision-Nest} use only diversity or diversity plus exploration fitness. Performance improves with exploration fitness. The rightmost group in the figures uses diversity, exploration, and the ad hoc fitness for GEESE-BT (green) and for GEESE with PPA structures (blue). The yellow box in the rightmost group uses diversity and exploration, and replaces the ad hoc and hand-tuned rewards with BT status. The rightmost group shows that replacing ad hoc fitness with BT status yields substantial improvement. Out of 64 simulation runs, BeTr-GEESE succeed 75\% of the time, whereas GEESE-BT succeeded only 9.3\%.

\section{Performance of Fixed Programs}
\label{sec:popquality}

\br-GEESE agents succeed while evolving, but they perform poorly once evolution stops (fixed agents). This section presents results for various mixtures of 100 fixed agents.

\subsection{Homogeneous Populations of Best-Performing Agents} 

Create a homogeneous population by selecting the fittest agent at the end of the evolution and forming a population with 100 copies of that agent. Homogeneous populations of evolved \br-GEESE and GEESE-BT agents fail on both tasks. The performance of homogeneous populations is poor because of the types of programs evolved. The foraging task requires at least four PPA sub-trees: explore the environment and find the site, carry the food, bring the food back to the hub, and drop the food at the hub. Among 100 independent evolution experiments, 3235 different \br-GEESE programs were evolved, 97.8\% had just one PPA sub-tree, and 2\% had two PPA sub-trees. These static programs are not capable of solving the problem by themselves. By contrast, GEESE-BT more frequently produced agents with all four necessary subtrees but still failed, indicating heterogeneous controllers are needed.

\subsection{Heterogeneous Populations of High-Performing Agents}
\label{subsec:sampling}

\begin{wrapfigure}{r}{.5\linewidth}
  \centering %
	\includegraphics[width=1\linewidth, keepaspectratio]{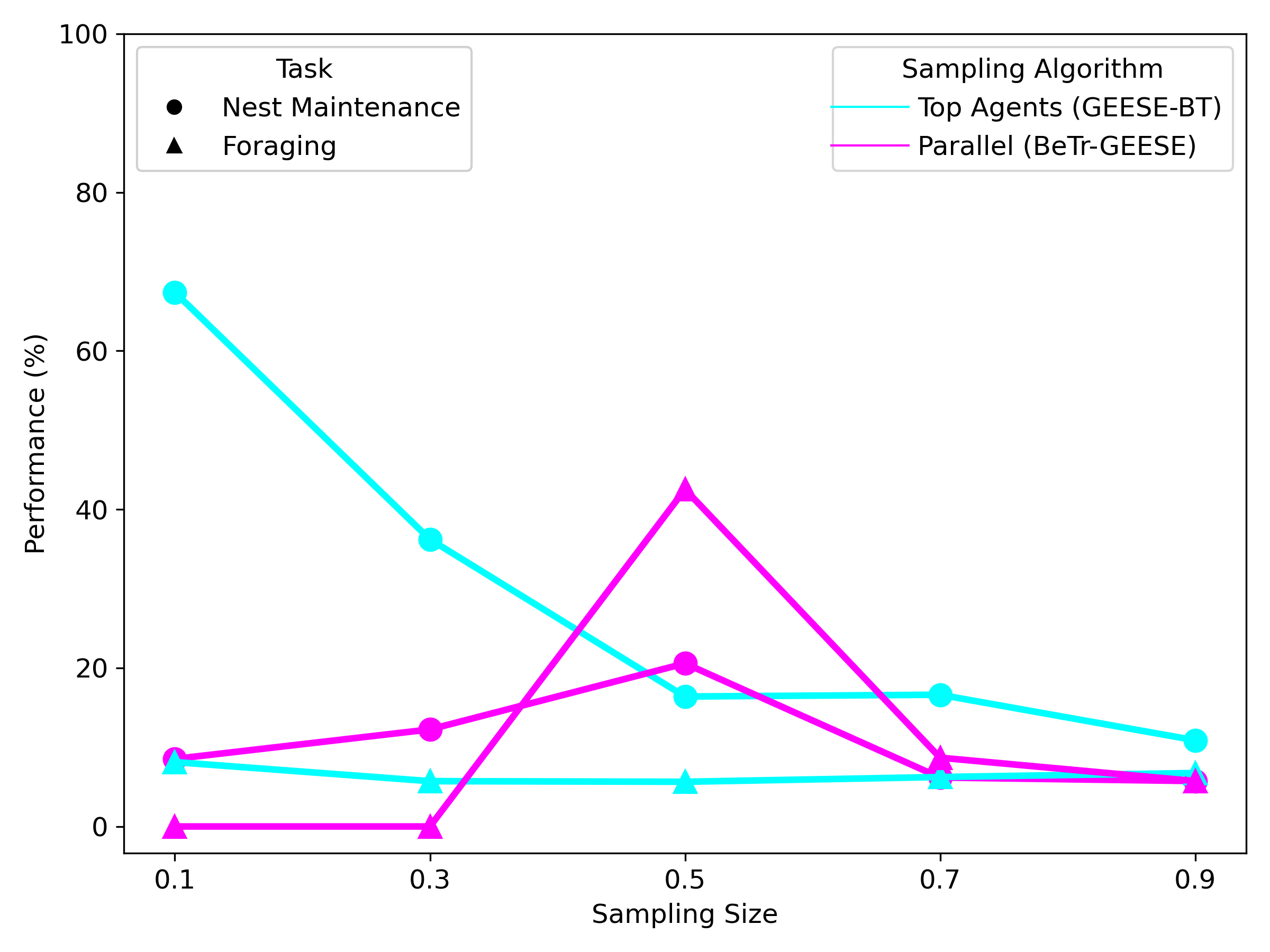}
    \caption{\footnotesize{Population quality for populations created by sampling the top $n\%$ of agents for GEESE-BT and remixing \br-GEESE agents.}}
    \label{fig:BeTr-GEESE_behavior_sample}
\end{wrapfigure}

The fitness of an individual agent can be deceiving because the agent might be fit only when other agents in a heterogeneous population are performing necessary supporting tasks~\citep{page2010diversity}. Form a heterogeneous population by sorting the agents at the end of evolution by their fitness value, identifying the top n\% of the agents, and then cloning those agents to create 100 agents. As with the homogeneous agents, the performance of a heterogeneous population of \br-GEESE agents was terrible; less than 1\% of the food available is brought to the hub so the results are not shown in the figure. Each agent is capable of only doing one subtask, and that means that agents cannot both pick up and drop objects. By contrast, Figure~\ref{fig:BeTr-GEESE_behavior_sample} shows that heterogenous blends of GEESE-BT agents often succeed, especially on the nest-maintenance task, precisely because they can learn more complicated programs. However, as more types of agents are added to the population, interference occurs, presumably because less fit agents perform only partial tasks and thus prevent top-performing agents from fully performing all tasks.

\subsection{Heterogeneous Populations of Blended Agents}
The better performance of the more complicated GEESE-BT programs suggests a way to blend the modular \br-GEESE programs so that they are more complicated as follows.  First, sort agents most fit to least fit. Second, select the top $n\%$ of the agents. Each of these agents typically has a BT that performs only one subtask. Third, {\tt or} these agents together by forming a root BT node with a parallel node and then adding agents as children to this root node.  The parallel control node was chosen as it loosely acts as a logical {\tt or}, meaning that the \textit{blended agent} allows its modular \br-GEESE programs (sub-trees) to contribute to the behavior of the agent. For each agent, the order of the sub-trees is randomized so that different agents try execute subtasks in different orders. 

Figure~\ref{fig:BeTr-GEESE_behavior_sample} shows performance of heterogeneous populations of \br-GEESE agents formed using this blended approach. Heterogeneous populations of blended GEESE-BT agents are not shown because they perform no better than the heterogeneous populations formed in Section~\ref{subsec:sampling}. The performance of blended \br-GEESE populations slowly increases and peaks at 0.5 and then decreases gradually for both tasks. Blending works precisely because the  resulting agents capture pieces of agents capable of successfully performing needed subtasks.

\section{Conclusion and Future Work}
The PPA-structures used in \br-GEESE cause agents to learn simple programs capable of performing required subtasks. The evolving population of \br-GEESE agents succeed because mutation, crossover, and lateral transfer cause the agents to change between simple programs while learning, effectually ``time multiplexing'' between the simple programs.   Time-multiplexing allows agents to adapt to different circumstances in their learning environment, effectually negating the need to rely on static strategies. 
GEESE-BT's success while learning is low because it is difficult to learn complicated programs even with the ad hoc reward structures designed to promote efficient learning. Heterogeneous populations of fixed \br-GEESE agents perform poorly because each agent learns only a portion of the entire task. The success of the heterogeneous population of can be improved by blending the simple agents together, but performance is still not satisfactory. 





Future work should explore \br-GEESE agents could rapidly adapt to changing conditions, provided that the grammar has a sufficiently rich set of execution nodes and primitive behaviors. Additionally, future work should explore genetic algorithm hyper-parameter settings that might enable \br-GEESE to evolve complex, resilient behaviors more efficiently.

\appendix
\section{Primitive Behaviors}
\label{section:PrimitiveBehaviors}
\textbf{Spatial Behaviors.} The \textit{CompositeDrop} behavior enables agent to drop the item it is currently carrying at the current location, and the \textit{CompositeSingleCarry} allows an agent to pickup an object at its current location. The \textit{MoveTowards} primitive behavior allows the agent to move a unit step towards a particular object in the environment, and \textit{MoveAway} moves the agent a unit step away from an object. \textit{Explore} moves the agent a unit step in a random direction.  Obstacle avoidance in \textit{MoveTowards}, \textit{MoveAway}, and \textit{Explore} use the simple bug following algorithm from~\cite{lumelsky1987path}, which checks for obstacles or traps in its line-of-sight. If an object is detected, the agent moves one unit distance parallel to the object's surface. 


\section{PPA Grammar}
\label{app:bnfgrammar}
\begin{footnotesize}
\begin{numberedgrammar}
\label{gram:swarm}
\squash      


<root> ::= <sequence> | <selector>

<sequence> ::= [Sequence]<ppa>[/Sequence]| [Sequence]<root><root>[/Sequence] \\
\ \ [Sequence]<sequence><root>[/Sequence]

<selector> ::= [Selector]<ppa>[/Selector]|[Selector]<root><root>[/Selector] \\
\ \ [Selector]<selector><root>[/Selector]

<ppa> ::= [Selector]<postconditions><ppasequence>[/Selector]

<postconditions> ::= <SuccessNode> | <ppa> |[Sequence]<postcondition>[/Sequence]

<postcondition> ::= <postcondition>[PostCnd]<postconditiont>\\
\ \ [/PostCnd]|[PostCnd]<postconditiont>[/PostCnd]

<postconditiont> ::= NeighbourObjects\_<objects>|...| IsVisitedBefore\_<sobjects>


<ppasequence> ::= [Sequence]<preconditions>[Act]<action>\\
\ \ [/Act][/Sequence]|[Sequence]<constraints>[Act]<action>[/Act][/Sequence]|\\
\ \ [Sequence]<preconditions><constraints> [Act]<action>[/Act][/Sequence]

<preconditions> ::= [Sequence]<precondition>[/Sequence]

<precondition> ::= <precondition>[PreCnd]<preconditiont> \\
\ \ [/PreCnd]| [PreCnd]<preconditiont>[/PreCnd]

<preconditiont> ::= IsDropable\_<sobjects>| ... | IsCarrying\_<dobjects>

<constraints> ::= [Sequence]<constraint>[/Sequence]

<constraint> ::= <constraint>[Cnstr]<constraintt>[/Cnstr]| [Cnstr]<constraintt>[/Cnstr]

<constraintt> ::= CanMove |...| IsDropable\_<sobjects>

<action> ::= MoveAway\_<sobjects>|...|Explore

<objects> ::= <sobjects>|<dobjects>

<sobjects> ::= Hub|Sites

<oobjects> ::= Obstacles|Trap

<dobjects> ::= Food|Debris

<SuccessNode> :: = [PostCnd]DummyNode[/PostCnd]
\end{numberedgrammar}
\end{footnotesize}


Productions 1--4 define control nodes as part of a hierarchical set of PPA-structured sub-trees. 
Productions 5--7 define postconditions, which can be dummy nodes, an embedded PPA sub-tree, or condition nodes. Production rule 20, the \textit{DummyNode}, is a condition node that always returns Success. 
Production~8 defines the right sub-tree of a PPA structure that sequentially combined preconditions and actions. Productions 9--11, define preconditions using condition nodes. 
Productions 12--14,  define constraints using condition nodes. 
Production~15 calls the BT-based subtrees in which the  primitive behaviors of the agent are defined.
Productions 16--18 define static elements in the swarm environment, specifically a hub, sites, obstacles, and traps, all of which have fixed sizes. Production~19 defines movable objects, food and debris, in the environment. A \textit{hub} is where agents originate, a \textit{site} is a source of food for the agents, an \textit{obstacle} obstructs the path of the agent and is not passable, a \textit{trap} kills the agent when it comes in contact, \textit{food} is an object to be carred to the hub, and \textit{debris} is an object to be removed from the hub.  All execution nodes have self-explanatory names.

\bibliography{iclr2022_conference}
\bibliographystyle{iclr2022_conference}

\end{document}

%% file: math_commands.tex
\usepackage{amsmath,amsfonts,bm}

\def\eqref#1{equation~\ref{#1}}

\def\1{\bm{1}}

\DeclareMathAlphabet{\mathsfit}{\encodingdefault}{\sfdefault}{m}{sl}
\SetMathAlphabet{\mathsfit}{bold}{\encodingdefault}{\sfdefault}{bx}{n}













%% file: iclr2022_conference.bbl
\begin{thebibliography}{44}
\providecommand{\natexlab}[1]{#1}
\providecommand{\url}[1]{\texttt{#1}}
\expandafter\ifx\csname urlstyle\endcsname\relax
  \providecommand{\doi}[1]{doi: #1}\else
  \providecommand{\doi}{doi: \begingroup \urlstyle{rm}\Url}\fi

\bibitem[Brooks(1986)]{brooks1986robust}
R.~Brooks.
\newblock A robust layered control system for a mobile robot.
\newblock \emph{IEEE Journal on Robotics and Automation}, 2\penalty0
  (1):\penalty0 14--23, 1986.

\bibitem[Cli et~al.(1993)Cli, Husbands, and Harvey]{cli1993evolving}
D.~Cli, P.~Husbands, and I.~Harvey.
\newblock Evolving visually guided robots.
\newblock In \emph{From Animals to Animats~2. Proc.~of the 2nd Intl.~Conf.~on
  Simulation of Adaptive Behavior}, pp.\  374--383. MIT Press, 1993.

\bibitem[Colledanchise \& {\"O}gren(2017)Colledanchise and
  {\"O}gren]{colledanchise2017behavior}
M.~Colledanchise and P.~{\"O}gren.
\newblock How behavior trees modularize hybrid control systems and generalize
  sequential behavior compositions, the subsumption architecture, and decision
  trees.
\newblock \emph{IEEE Transactions on Robotics}, 33\penalty0 (2):\penalty0
  372--389, 2017.

\bibitem[Colledanchise \& {\"O}gren(2018)Colledanchise and
  {\"O}gren]{colledanchise2018behavior}
M.~Colledanchise and P.~{\"O}gren.
\newblock Behavior trees in robotics and al: An introduction.
\newblock 2018.

\bibitem[Colledanchise et~al.(2019)Colledanchise, Almeida, and
  {\"O}gren]{colledanchise2019towards}
M.~Colledanchise, D.~Almeida, and P.~{\"O}gren.
\newblock Towards blended reactive planning and acting using behavior trees.
\newblock In \emph{2019 ICRA}, pp.\  8839--8845. IEEE, 2019.

\bibitem[Dorigo et~al.(2006)Dorigo, Birattari, and Stutzle]{dorigo2006ant}
M.~Dorigo, M.~Birattari, and T.~Stutzle.
\newblock Ant colony optimization.
\newblock \emph{IEEE Computational Intelligence Magazine}, 1\penalty0
  (4):\penalty0 28--39, 2006.

\bibitem[Duarte et~al.(2016)Duarte, Costa, Gomes, Rodrigues, Silva, Oliveira,
  and Christensen]{duarte2016evolution}
M.~Duarte, V.~Costa, J.~Gomes, T.~Rodrigues, F.~Silva, S.~M. Oliveira, and
  A.~L. Christensen.
\newblock Evolution of collective behaviors for a real swarm of aquatic surface
  robots.
\newblock \emph{PloS One}, 11\penalty0 (3):\penalty0 e0151834, 2016.

\bibitem[Engebr{\aa}ten et~al.(2018)Engebr{\aa}ten, Moen, Yakimenko, and
  Glette]{engebraaten2018evolving}
Sondre~A Engebr{\aa}ten, Jonas Moen, Oleg Yakimenko, and Kyrre Glette.
\newblock Evolving a repertoire of controllers for a multi-function swarm.
\newblock In \emph{International Conference on the Applications of Evolutionary
  Computation}, pp.\  734--749. Springer, 2018.

\bibitem[Fan et~al.(2021)Fan, Xiong, Li, and Wang]{fan2021interpretability}
F.~Fan, J.~Xiong, M.~Li, and G.~Wang.
\newblock On interpretability of artificial neural networks: A survey.
\newblock \emph{IEEE Transactions on Radiation and Plasma Medical Sciences},
  2021.

\bibitem[Fenton et~al.(2017)Fenton, McDermott, Fagan, Forstenlechner, Hemberg,
  and O'Neill]{fenton2017ponyge2}
Michael Fenton, James McDermott, David Fagan, Stefan Forstenlechner, Erik
  Hemberg, and Michael O'Neill.
\newblock Ponyge2: Grammatical evolution in python.
\newblock In \emph{Proceedings of the Genetic and Evolutionary Computation
  Conference Companion}, pp.\  1194--1201, 2017.

\bibitem[Ferrante et~al.(2013)Ferrante, Du{\'e}{\~n}ez-Guzm{\'a}n, Turgut, and
  Wenseleers]{ferrante2013geswarm}
E.~Ferrante, E.~Du{\'e}{\~n}ez-Guzm{\'a}n, A.~E. Turgut, and T.~Wenseleers.
\newblock Geswarm: Grammatical evolution for the automatic synthesis of
  collective behaviors in swarm robotics.
\newblock In \emph{Proc.~ of the 15th annual GECCO conference}, pp.\  17--24.
  ACM, 2013.

\bibitem[Fikes \& Nilsson(1971)Fikes and Nilsson]{fikes1971strips}
R.~E. Fikes and N.~J. Nilsson.
\newblock Strips: A new approach to the application of theorem proving to
  problem solving.
\newblock \emph{Artificial intelligence}, 2\penalty0 (3-4):\penalty0 189--208,
  1971.

\bibitem[Gordon(2010)]{gordon2010ant}
D.~M. Gordon.
\newblock \emph{Ant encounters: interaction networks and colony behavior},
  volume~1.
\newblock Princeton University Press, 2010.

\bibitem[Jablonka \& Lamb(2014)Jablonka and Lamb]{jablonka2014evolution}
Eva Jablonka and Marion~J Lamb.
\newblock \emph{Evolution in four dimensions, revised edition: {G}enetic,
  epigenetic, behavioral, and symbolic variation in the history of life}.
\newblock MIT press, 2014.

\bibitem[Karaboga \& Basturk(2007)Karaboga and Basturk]{karaboga2007powerful}
D.~Karaboga and B.~Basturk.
\newblock A powerful and efficient algorithm for numerical function
  optimization: artificial bee colony (abc) algorithm.
\newblock \emph{Journal of global optimization}, 39\penalty0 (3):\penalty0
  459--471, 2007.

\bibitem[Kazil et~al.(2020)Kazil, Masad, and Crooks]{python-mesa-2020}
Jackie Kazil, David Masad, and Andrew Crooks.
\newblock Utilizing python for agent-based modeling: The mesa framework.
\newblock In Robert Thomson, Halil Bisgin, Christopher Dancy, Ayaz Hyder, and
  Muhammad Hussain (eds.), \emph{Social, Cultural, and Behavioral Modeling},
  pp.\  308--317, Cham, 2020. Springer International Publishing.
\newblock ISBN 978-3-030-61255-9.

\bibitem[Knoblock(1995)]{knoblock1995planning}
C.~A. Knoblock.
\newblock Planning, executing, sensing, and replanning for information
  gathering.
\newblock Technical report, University of Southern California, 1995.

\bibitem[K{\"o}nig et~al.(2009)K{\"o}nig, Mostaghim, and
  Schmeck]{konig2009decentralized}
L.~K{\"o}nig, S.~Mostaghim, and H.~Schmeck.
\newblock Decentralized evolution of robotic behavior using finite state
  machines.
\newblock \emph{International Journal of Intelligent Computing and
  Cybernetics}, 2\penalty0 (4):\penalty0 695--723, 2009.

\bibitem[Kriesel et~al.(2008)Kriesel, Cheung, Sitti, and
  Lipson]{kriesel2008beanbag}
D.~M. Kriesel, E.~Cheung, M.~Sitti, and H.~Lipson.
\newblock Beanbag robotics: Robotic swarms with 1-dof units.
\newblock In \emph{ANTS 2008}, pp.\  267--274. Springer, 2008.

\bibitem[Kucking et~al.(2018)Kucking, Ligot, Bozhinoski,
  et~al.]{jonas2018behavior}
J.~Kucking, A.~Ligot, D.~Bozhinoski, et~al.
\newblock Behavior trees as a control architecture in the automatic design of
  robot swarms.
\newblock In \emph{ANTS 2018}. IEEE, 2018.

\bibitem[Kuckling et~al.(2021)Kuckling, Van~P., and
  Birattari]{kuckling2021automatic}
J.~Kuckling, Vincent Van~P., and M.~Birattari.
\newblock Automatic modular design of behavior trees for robot swarms with
  communication capabilites.
\newblock In \emph{EvoApplications}, pp.\  130--145, 2021.

\bibitem[Lane(2015)]{lane2015vital}
Nick Lane.
\newblock \emph{The vital question: {E}nergy, evolution, and the origins of
  complex life}.
\newblock WW Norton \& Company, 2015.

\bibitem[Lee(1999)]{lee1999evolving}
Wei-Po Lee.
\newblock Evolving complex robot behaviors.
\newblock \emph{Information Sciences}, 121\penalty0 (1-2):\penalty0 1--25,
  1999.

\bibitem[Lewis(1996)]{lewis1996neural}
F.~L. Lewis.
\newblock Neural network control of robot manipulators.
\newblock \emph{IEEE Expert}, 11\penalty0 (3):\penalty0 64--75, 1996.

\bibitem[Lewis et~al.(1992)Lewis, Fagg, and Solidum]{lewis1992genetic}
M.~A. Lewis, A.~H. Fagg, and A.~Solidum.
\newblock Genetic programming approach to the construction of a neural network
  for control of a walking robot.
\newblock In \emph{Robotics and Automation, 1992. Proceedings., 1992 IEEE
  Intl.~Conf.~on}, pp.\  2618--2623. IEEE, 1992.

\bibitem[Lumelsky \& Stepanov(1987)Lumelsky and Stepanov]{lumelsky1987path}
V.~J. Lumelsky and A.~A. Stepanov.
\newblock Path-planning strategies for a point mobile automaton moving amidst
  unknown obstacles of arbitrary shape.
\newblock \emph{Algorithmica}, 2\penalty0 (1):\penalty0 403--430, 1987.

\bibitem[Nelson et~al.(2009)Nelson, Barlow, and Doitsidis]{nelson2009fitness}
Andrew~L Nelson, Gregory~J Barlow, and Lefteris Doitsidis.
\newblock Fitness functions in evolutionary robotics: A survey and analysis.
\newblock \emph{Robotics and Autonomous Systems}, 57\penalty0 (4):\penalty0
  345--370, 2009.

\bibitem[Neupane \& Goodrich(2019{\natexlab{a}})Neupane and
  Goodrich]{neupane2019designing}
A.~Neupane and M.~A. Goodrich.
\newblock Designing emergent swarm behaviors using behavior trees and
  grammatical evolution.
\newblock In \emph{Proc. of the 18th AAMAS conference}, pp.\  2138--2140,
  2019{\natexlab{a}}.

\bibitem[Neupane \& Goodrich(2019{\natexlab{b}})Neupane and
  Goodrich]{neupane2019learning}
A.~Neupane and M.~A. Goodrich.
\newblock Learning swarm behaviors using grammatical evolution and behavior
  trees.
\newblock In \emph{IJCAI}, pp.\  513--520, 2019{\natexlab{b}}.

\bibitem[Neupane et~al.(2018)Neupane, Goodrich, and Mercer]{neupane2018geese}
A.~Neupane, M.~A. Goodrich, and E.~G. Mercer.
\newblock Geese: grammatical evolution algorithm for evolution of swarm
  behaviors.
\newblock In \emph{Proc.~ of the 20th annual GECCO conference}, pp.\
  999--1006, 2018.

\bibitem[{\"O}gren(2020)]{ogren2020convergence}
P.~{\"O}gren.
\newblock Convergence analysis of hybrid control systems in the form of
  backward chained behavior trees.
\newblock \emph{IEEE Robotics and Automation Letters}, 5\penalty0 (4):\penalty0
  6073--6080, 2020.

\bibitem[O'Neill \& Ryan(2001)O'Neill and Ryan]{o2001grammatical}
M.~O'Neill and C.~Ryan.
\newblock Grammatical evolution.
\newblock \emph{IEEE Transactions on Evolutionary Computation}, 5\penalty0
  (4):\penalty0 349--358, 2001.

\bibitem[Page(2010)]{page2010diversity}
S.~E. Page.
\newblock \emph{Diversity and Complexity}.
\newblock Princeton University Press, 2010.

\bibitem[Parashar et~al.(2021)Parashar, Naik, Hu, and
  Christensen]{parashar2021meta}
P.~Parashar, A.~Naik, J.~Hu, and H.~I Christensen.
\newblock Meta-modeling of assembly contingencies and planning for repair.
\newblock \emph{arXiv preprint arXiv:2103.07544}, 2021.

\bibitem[Petrovic(2008)]{petrovic2008evolving}
Pavel Petrovic.
\newblock Evolving behavior coordination for mobile robots using distributed
  finite-state automata.
\newblock In \emph{Frontiers in evolutionary robotics}. InTech, 2008.

\bibitem[Pint{\'e}r-Bartha et~al.(2012)Pint{\'e}r-Bartha, Sobe, and
  Elmenreich]{pinter2012towards}
A.~Pint{\'e}r-Bartha, A.~Sobe, and W.~Elmenreich.
\newblock Towards the light—comparing evolved neural network controllers and
  finite state machine controllers.
\newblock In \emph{Proc. of the 10th Workshop on WISES}, pp.\  83--87. IEEE,
  2012.

\bibitem[Quammen(2018)]{quammen2018tangled}
David Quammen.
\newblock \emph{The tangled tree: {A} radical new history of life}.
\newblock Simon and Schuster, 2018.

\bibitem[Sprague et~al.(2018)Sprague, {\"O}zkahraman, Munafo, Marlow,
  et~al.]{sprague2018improving}
C.~I. Sprague, {\"O}.~{\"O}zkahraman, A.~Munafo, R.~Marlow, et~al.
\newblock Improving the modularity of auv control systems using behaviour
  trees.
\newblock In \emph{2018 IEEE/OES Autonomous Underwater Vehicle Workshop (AUV)},
  pp.\  1--6. IEEE, 2018.

\bibitem[Steiner(1972)]{steiner1972group}
Ivan~Dale Steiner.
\newblock \emph{Group process and productivity}.
\newblock Academic press, 1972.

\bibitem[Stonier \& Staniaszek(2021)Stonier and Staniaszek]{py_trees}
Daniel Stonier and Michal Staniaszek.
\newblock {Behavior Tree implementation in Pyton}, 12 2021.
\newblock URL \url{https://github.com/splintered-reality/py_trees/}.

\bibitem[Sumpter \& Pratt(2003)Sumpter and Pratt]{sumpter2003modelling}
D.~Sumpter and S.~Pratt.
\newblock A modelling framework for understanding social insect foraging.
\newblock \emph{Behavioral Ecology and Sociobiology}, 53\penalty0 (3):\penalty0
  131--144, 2003.

\bibitem[Sutton(1984)]{sutton1984temporal}
R.~S. Sutton.
\newblock \emph{Temporal credit assignment in reinforcement learning}.
\newblock PhD thesis, University of Massachusetts Amherst, 1984.

\bibitem[Trianni et~al.(2003)Trianni, Gro{\ss}, Labella, {\c{S}}ahin, and
  Dorigo]{trianni2003evolving}
Vito Trianni, Roderich Gro{\ss}, Thomas~H Labella, Erol {\c{S}}ahin, and Marco
  Dorigo.
\newblock Evolving aggregation behaviors in a swarm of robots.
\newblock In \emph{European Conference on Artificial Life}, pp.\  865--874.
  Springer, 2003.

\bibitem[Valmari(1996)]{valmari1996state}
A.~Valmari.
\newblock The state explosion problem.
\newblock In \emph{Advanced Course on Petri Nets}, pp.\  429--528. Springer,
  1996.

\end{thebibliography}
